\documentclass[sigconf]{acmart}

\usepackage{multirow}
\usepackage{float}
\usepackage{hyperref}

\def\eg{\emph{e.g., }} 
\def\ie{\emph{i.e., }}

\AtBeginDocument{%
  }

\copyrightyear{2025}
\acmYear{2025}
\setcopyright{cc}
\setcctype{by-nc}
\acmConference[MM '25]{Proceedings of the 33rd ACM International Conference
on Multimedia}{October 27--31, 2025}{Dublin, Ireland}
\acmBooktitle{Proceedings of the 33rd ACM International Conference on
Multimedia (MM '25), October 27--31, 2025, Dublin, Ireland}
\acmDOI{10.1145/3746027.3762073}
\acmISBN{979-8-4007-2035-2/2025/10}

\begin{document}

\title{ERR@HRI 2.0 Challenge: Multimodal Detection of Errors and Failures in Human-Robot Conversations}

\author{Shiye Cao}
\affiliation{%
  \institution{Johns Hopkins University}
  \city{Baltimore}
  \country{USA}}
\email{shiyecao@cs.jhu.edu}

\author{Maia Stiber}
\affiliation{%
  \institution{Microsoft Research}
  \city{Redmond}
  \country{USA}}
\email{maiastiber@microsoft.com}

\author{Amama Mahmood}
\affiliation{%
  \institution{Johns Hopkins University}
  \city{Baltimore}
  \country{USA}}
\email{amama.mahmood@jhu.edu}

\author{Maria Teresa Parreira}
\affiliation{%
  \institution{Cornell University}
  \city{Ithaca}
  \country{USA}}
\email{mb2554@cornell.edu}

\author{Wendy Ju}
\affiliation{%
  \institution{Cornell Tech}
  \city{New York}
  \country{USA}}
\email{wendyju@cornell.edu}

\author{Micol Spitale}
\affiliation{%
  \institution{Politecnico di Milano}
  \city{Milano}
  \country{Italy}}
\email{micol.spitale@polimi.it}

\author{Hatice Gunes}
\affiliation{%
  \institution{University of Cambridge}
  \city{Cambridge}
  \country{UK}}
\email{hatice.gunes@cl.cam.ac.uk}

\author{Chien-Ming Huang}
\affiliation{%
  \institution{Johns Hopkins University}
  \city{Baltimore}
  \country{USA}}
\email{chienming.huang@jhu.edu}

\renewcommand{\shortauthors}{Shiye Cao et al.}

\begin{abstract}
The integration of large language models (LLMs) into conversational robots has made human-robot conversations more dynamic. Yet, LLM-powered conversational robots remain prone to errors, \eg misunderstanding user intent, prematurely interrupting users, or failing to respond altogether. Detecting and addressing these failures is critical for preventing conversational breakdowns, avoiding task disruptions, and sustaining user trust. To tackle this problem, the ERR@HRI 2.0 Challenge provides a multimodal dataset of LLM-powered conversational robot failures during human-robot conversations and encourages researchers to benchmark machine learning models designed to detect robot failures. The dataset includes 16 hours of dyadic human-robot interactions, incorporating facial, speech, and head movement features. Each interaction is annotated with
the presence or absence of robot errors from the system perspective, and perceived user intention to correct for a mismatch between robot behavior and user expectation. 
Participants are invited to form teams and develop machine learning models that detect these failures using multimodal data. Submissions will be evaluated using various performance metrics, including detection accuracy and false positive rate.
This challenge represents another key step toward improving failure detection in human-robot interaction through social signal analysis.
\end{abstract}






\maketitle

\begin{figure}[th]
  \includegraphics[]{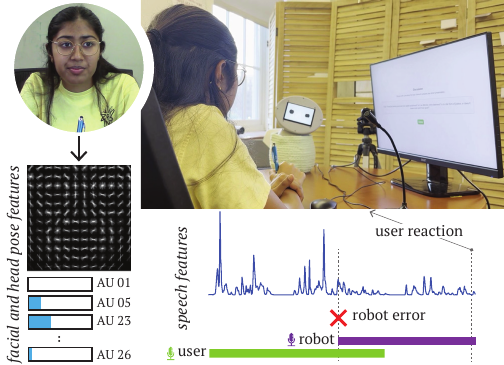}
  \caption{Illustration of the interaction and data recording setup, together with extracted features.}
  %
  \label{fig:dataset}
\end{figure}

\section{Introduction}
In recent years, the integration of large language models (LLMs) into conversational robots---embodied media---has enabled more natural human-robot conversations. While LLMs can mitigate certain speech recognition errors that often cause traditional conversational robots to fail, these advanced systems remain vulnerable to issues such as interrupting users and misinterpreting user intent \cite{mahmood2025user, mahmood2024situated, cao2025interruption}. Such failures can lead to conversational breakdowns, disrupt interaction flow, and negatively affect users' perception of the robot. As LLM-powered conversational robots are deployed in real-world scenarios, the need to understand and mitigate their failures is a pressing issue. Prior work has discovered some strategies to help robots rebuild user trust and recover from errors~\cite{karli2023if, esterwood2021you, AxelssonEtAl-HRI2024}; however, the application of these strategies hinges on the robot's ability to detect its own errors effectively~\cite{stiber2023using}. Status quo robot error detection methods often either depend on task-specific or domain-specific information---which can be hard to generalize across people, tasks, and errors---or rely on users to detect and report errors---which can be delayed. Recent works showed the possibility of detecting robot error through leveraging social signals exhibited by user reactions and responses to robot errors in multimedia streams \cite{stiber2025robot, bremers2023using, spitale2023longitudinal,pramanick2024prisca,janssens2024predicting,wachowiak2024time}. To further explore this potential and benchmark multimodal robot failure detection in HRI, the ERR@HRI 2.0 challenge focuses on \textbf{multimodal detection of robot errors and failures in dyadic human-robot conversations}. 

The ERR@HRI initiative serves as a platform for benchmarking not only HRI datasets but also machine learning models aimed at detecting robot errors and failures from multimedia. The 2024 ERR@HRI challenge, held as part of the 2024 ACM International Conference on Multimodal Interaction (ICMI'24), focused on modeling robot errors using a multimodal dataset collected in a real-world setting where a robotic coach delivered well-being coaching practices to each individual separately over four weeks \cite{spitale2024err}. Building on this, the ERR@HRI 2.0 challenge uses a new multimodal dataset (see Fig. \ref{fig:dataset}) comprised of multimodal behavioral signals extracted from audio and video recordings of human conversations with two distinctive LLM-powered embodied media: 
a social robot with more anthropomorphic features and expressions, and a smart speaker. The user and the robot collaborated on five tasks during the conversations, ranging from medical self-diagnosis, trip planning, selecting items more beneficial during a desert survival simulation, to discussing whether the federal government should ban capital punishment and whether universities should have their own police force. Additionally, the ERR@HRI 2.0 challenge is also unique in that it contains two sub-challenges to tackle robot error detection from two perspectives. Sub-challenge 1 targets the detection of robot errors and failures from the system perspective, while sub-challenge 2 targets the detection of robot errors and failures from the user perspective. 

Through this challenge, we hope to bring together researchers and practitioners from the multimedia, robotics, and human-robot interaction communities, while highlighting the underlying multimedia aspects of the problem. 
The techniques developed through this challenge will contribute to advancing our understanding of how multimedia can enhance the analysis and interpretation of interactions between humans and autonomous robots. 

\section{Related Work}
\label{sec:related-work}
\subsection{Robot Errors}
As robots become more capable, they are increasingly being integrated into daily life, assisting and collaborating with human users across a range of tasks and contexts~\cite{mahmood2025voice, gerling2016robot, portugal2019study, scassellati2018improving}. However, errors and failures remain common and often inevitable in robots deployed in real-world settings~\cite{carlson2005ugvs, honig2018understanding, mirnig2015impact}. For a robot to be truly effective in a human-robot team, it must not only complete its assigned tasks but also adapt to the user and align its behavior with the user's expectations---thereby enhancing team performance and user satisfaction~\cite{tabrez2020survey}. This highlights a limitation in existing robot evaluation approaches that only assess the robot based on task completion. Instead, we need to consider how the task was performed and whether the robot's behavior conformed to their human teammate's expectations. Therefore, in this work, we adopt a broader definition of robot errors to assess the robot behaviors from two distinct perspectives: 1) \textit{robot errors from the system perspective}---deviations in the robot behavior from its expected behavior by system design; and 2) \textit{robot errors from the user perspective}---deviations in the robot behavior from the user's mental model of the expected robot behavior. 

\subsection{Robot Error Detection Using Social Signals}
Social signals are multimodal behavioral cues (both verbal and non-verbal) that can convey information about one's emotions, intentions, attitudes, and social dynamics within interactions \cite{vinciarelli2008social, cao2022understanding, candon2023nonverbal, sauppe2014social, stiber2024uh}. In human-robot interaction (HRI), users' natural responses (social signals) during the interaction can impart information about their own internal state (\eg user uncertainty) and their mental model of the robot \cite{vinciarelli2011bridging, tabrez2020survey, stiber2022modeling, scherf2024you, huang2013repertoire}. As such, social signals exhibited by users have been used to aid collaboration in HRI by conveying user preferences (\eg~\cite{candon2023nonverbal}), user's need for help (\eg{\cite{wilson2022help}}), and engagement breakdowns (\eg{\cite{ben2019early}}). 

Social signals have also been used to facilitate automatic error detection in HRI. Robot errors naturally elicit responses from users~\cite{mirnig2015impact}. Capitalizing on those elicitations expressed as multimodal streams, recent works have shown the possibility of using them for detecting robot errors~\cite{stiber2022modeling, stiber2023using, stiber2025robot, bremers2023using, spitale2023longitudinal, kontogiorgos2021systematic, giuliani2015systematic, spitale2024err}. The 2024 ERR@HRI challenge~\cite{spitale2024err} at the 2024 ACM International Conference on Multimodal Interaction focused on using a multimodal dataset collected in a real-world setting where a robotic coach delivered well-being coaching practices to each individual separately over four weeks. In an effort to further explore the potential of this and to benchmark multimodal robot failure detection in HRI, the ERR@HRI 2.0 challenge focuses on the multimodal detection of robot errors and failures in dyadic human-robot conversations across two unique embodiments and five tasks.

\begin{table*}[t]
\caption{Dataset characteristics}
\begin{tabular}{c|c|c|c|cc}
\multirow{2}{*}{\textbf{Embodiments}}     & \multirow{2}{*}{\textbf{Tasks}} & \multirow{2}{*}{\textbf{Time}} & \multirow{2}{*}{\textbf{\# of Sessions}} & \multicolumn{2}{c}{\textbf{\# of Robot Errors}}                     \\ \cline{5-6} 
                                 &                        &                       &                                 & \multicolumn{1}{c|}{\textbf{System Perspective}} & \textbf{User Perspective} \\ \hline
\multirow{2}{*}{\textbf{Social Robot \cite{cao2025interruption}}}    & Survival               & 2 hr                  & 22                              & \multicolumn{1}{c|}{73}                 & 29               \\ \cline{2-6} 
                                 & Discussion             & 1 hr 43 min           & 20                              & \multicolumn{1}{c|}{85}                 & 82               \\ \hline
\multirow{3}{*}{\textbf{Voice Assistant\cite{mahmood2025user}}} & Medical Self-Diagnosis & 3 hr 50 min           & 20                              & \multicolumn{1}{c|}{72}                 & 40               \\ \cline{2-6} 
                                 & Trip Planning          & 4 hr 41 min           & 20                              & \multicolumn{1}{c|}{96}                 & 46               \\ \cline{2-6} 
                                 & Discussion             & 3 hr 36 min           & 19                              & \multicolumn{1}{c|}{88}                 & 27               \\ \hline
\multicolumn{1}{l|}{}            & \textbf{Total}                  & \textbf{15 hr 50 min}          & \textbf{101}                             & \multicolumn{1}{c|}{\textbf{414}}                & \textbf{224}          
\end{tabular}
\label{tab:dataset}
\end{table*}

\section{The ERR@HRI 2.0 Challenge}
In continuation of previous year's ERR@HRI~\cite{spitale2024err} efforts to encourage researchers to develop and benchmark multimodal machine learning models to detect robot errors during human-robot interaction, the ERR@HRI 2.0 challenge invites participants to develop multimodal ML models designed to detect conversational robot failures using a dataset that includes facial, head pose, and speech features extracted from 16 hours of dyadic human-robot conversational media.

\subsection{Definition of Robot Errors}
In this work, we are interested in detecting robot errors both from the system perspective and the user perspective (Section~\ref{sec:related-work}). These two types of robot errors overlap but are distinctive as deviations from the system's designed behavior may not always constitute errors from the user's perspective. Similarly, system's designed behavior that deviates from user's expected system behaviors are not always errors from the system's perspective but are considered errors or failures from the user's perspective. Therefore, it is important to consider both user perceived errors and system perceived errors when creating models that detect errors during human-robot interaction. We operationalize \emph{robot errors from the system perspective} as a noticable misalignment between robot behavior and its expected behavior by system design. We operationalized \emph{robot errors from the user perspective} as noticable verbal and non-verbal user-initiated disruptive interruptions that are intended to disrupt and correct for a misalignment between the robot behavior and the user's expected robot behavior. We want to emphasize that our operationalized definition of robot error used in this work does not capture all robot errors from the system perspective nor the user perspective. However, this definition allows for robot errors to be labeled from a third-party perspective, without having the participants
self-report when they felt like the robot made a mistake. A similar approach was used in prior work to define other user internal states, \eg confusion \cite{stiber2024uh}. 

\subsection{The Tasks}
Hence, the ERR@HRI 2.0 Challenge consists of two sub-challenges:
\begin{enumerate}
    \item Sub-challenge 1. Detection of robot errors from the system’s perspective
    \item Sub-challenge 2. Detection of robot errors from the user’s perspective
\end{enumerate}

In sub-challenge 1, we operationalize robot error from the system's perspective as deviations in the robot behavior from its designed behavior. Examples of this type of robot errors include user intention recognition errors, interrupting the user, and not responding to the user. 

In sub-challenge 2, we operationalize robot error from the user's perspective as detection of user intention to correct for a mismatch between robot behavior and their expectation (\ie user-initiated disruptive interruptions). Examples of this include user-initiated verbal and non-verbal interruptions that are perceived to be disruptive in intention. 

\subsection{Dataset}
The dataset consists of data from 42 users (21 female, 21 male), in a total of 101 sessions and roughly 950 minutes (15 hours and 50 minutes 48 seconds) of interaction collected in a previous dyadic human-social robot conversation study~\cite{cao2025interruption} and dyadic human-voice assistant conversation study~\cite{mahmood2025user}. Table~\ref{tab:dataset} summarizes additional characteristics of the dataset. 

The human-social robot conversational  data consists of 3 hours and 43 minutes 50 seconds of audio and video data of 22 unique participants interacting with a social robot to complete a set of one to two tasks including selected items to aid survival in a desert survival simulation (1 hour 43 minutes 11 seconds) and discussing whether the federal government should ban capital punishment (2 hours 39 seconds). For more details on the task and the study condition under which the data was collected, please refer to~\cite{cao2025interruption}. 

The human-voice assistant conversational data consists of 12 hours and 6 minutes 58 seconds of audio and video data, collected from 20 unique participants interacting with the voice assistant on a set of one to three tasks including medical self-diagnosis (3 hours 50 minutes 29 seconds), planning a day trip in Edinburgh (4 hours 40 minutes 45 seconds), and discussing whether universities should have their own police forces (3 hours 35 minutes 44 seconds). For more details on the task and the study condition under which the data was collected, please refer to the paper~\cite{mahmood2025user}.

\subsubsection{Feature Extraction}
We took the audio (user speech and robot speech) and video (camera facing user faces) recordings of the interactions from the two studies and used off-the-shelf state-of-the-art methods to extract facial features, head pose features, audio features, and transcribed speech features:

\begin{enumerate}
    \item Facial and head pose features: We used the OpenFace 2.2.0 toolkit~\cite{baltrusaitis2018openface} to identify facial expressions by detecting the presence of 18 facial action units (AUs) and estimating the intensity levels for 17 of those AUs. We also use the OpenFace 2.2.0 toolkit to estimate the location (x, y, and z coordinates) and rotation (x, y, and z coordinates) of the user's head. These features combined with the toolkit output on its confidence and success results in a total of 43 features per frame at a rate of 30 frames per second. 
    
    \item Audio features: We used the openSMILE toolbox~\cite{eyben2010opensmile} to extract 25 audio features from frames of 20ms length every 10ms using the eGeMAPSv02 feature set. The eGeMAPSv02 feature set contains loudness, alpha ratio, Hammarberg index, spectral slope between 0 and 500 Hz, spectral slope between 500 and 1500 Hz, spectral flux, first four Mel Frequency Cepstral Coefficients (MFCCs 1--4), F0, jitter local, shimmer local, harmonics-to-noise ratio, log energy ratio of F0,  frquency, bandwidth, and log amplitude relative to F0 of F1, F2, and F3.
    
    \item Transcribed speech features: We used Google's speaker diarization to detect user and robot speaker turns in audio recordings. Based on the result, we extracted 515 features showing when the robot/user is speaking, the word count of their speech content, the speech-to-text confidence, and created a 512-dimension embedding of the transcribed speech using a ViT-B/16 Transformer architecture Contrastive Language–Image Pre-training (CLIP)~\cite{radford2021learning} embedding. 
\end{enumerate}

To preserve user privacy, we only released the 581 multimodal behavioral features extracted from the recordings for modeling. 

\subsubsection{Labels}
Two coders annotated the videos using the Datavyu video annotation tool. The coders first familarized themselves with system design and expected behavior of the robot by design. Then, they marked the start and end time of observable deviation of robot behavior from its designed behavior, start and end time of any observable user reactions to the robot error from the system behavior (if any), and the start and end time of any observable user verbal and non-verbal disruptive interruption intended to correct for a mismatch between robot behavior and their expectation. The two coders first indepentently annotated ten sessions (two sessions from each task). Then, they discussed to resolve conflicts in the annotations from the ten sessions. They reached $100\%$ agreement after the discussion and proceeded to annotate the rest of the data. 

The labels were defined as follows: 
\begin{enumerate}
    \item \textbf{Robot error from system perspective}: The robot makes a mistake such as interrupting or not responding to the user, or responding with an error message or an utterance that is not appropriate for what the user has just said.    
    \item \textbf{Reaction to robot error}: Observable verbal and non-verbal user reaction to the \emph{robot error from the system perspective}. Users did not react to every error. 
    \item  \textbf{Robot error from user perspective}: The user displays behavior (verbal or non-verbal) that signals an intention to correct for a mismatch between robot behavior and their expectation such as user-initiated disruptive interruption. Disruptive interruption is defined as when the listener challenges the speaker’s control and disrupts the conversational flow to take the floor, change the subject, or avoid unwanted information. This behavior suggests that there exists some mismatch between the user's expectation for the robot and the robot's behavior.  
\end{enumerate}

We added annotations for observable user reaction to robot error from the system perspective as an additional potential label to faciliate training models to detect robot errors from the system perspective for sub-challenge 1. The dataset contained a total of 414 errors from the system perspective and 224 errors from the user perspective. See Table~\ref{tab:dataset} for the distribution of the labels across the different tasks. 

\subsubsection{Training and Test Sets}
We divided the ERR@HRI 2.0 dataset into a training set and test sets by splitting the dataset using a subject-independent strategy (\ie the training, validation, and testing sets do not include data from the same subjects). We randomly selected 20 sessions from 5 users to form the test set. This resulted in a training set composed of $81$ interactions ($47$ with voice assistants and $34$ with social robots); a test set composed of $20$ interactions ($12$ with voice assistants and $8$ with social robots). 

\subsection{Metrics}
We assess the model's performance on each of the two sub-challenges in two ways: 1) the model's error detection performance over fixed, pre-segmented windows, a simplified \emph{offline} evaluation scheme compared to the streaming scenario; 2) the model's error detection performance \emph{on-the-fly}, simulating how the model would perform in a streaming scenario. We use different metrics for \emph{offline} evaluation and \emph{on-the-fly} evaluation due to the difference in evaluation goals. 

We define an error as detected when the window in which the error is predicted overlaps with the ground truth error duration labeled. Following this definition, a prediction is a \textit{true positive} when it overlaps with the ground truth error/reaction windows. We added a tolerance of one second to the start and end labels to account for label inaccuracies. If the two windows overlap by any amount then that is considered a true positive. Following this logic, we consider a prediction to be a \textit{false positive}, when the prediction window does not overlap with any ground truth windows. We used these definitions to evaluate overall model performance. 

In summary, we used the following metrics to evaluate the trained models in this work: 

\noindent\textbf{\emph{Offline} evaluation metrics}:
\begin{itemize}
    \item Area Under the Receiver Operating Characteristic Curve (AUC)
    \item F1 score (F1) 
    \item Accuracy (Acc)
    \item Balanced Accuracy Score (UAR) \cite{brodersen2010balanced}
\end{itemize}

\noindent\textbf{\emph{On-the-fly} evaluation metrics}:
\begin{itemize}
    \item Percentage of errors/reactions detected (\%Detected): This metric evaluates the detection performance of the models. It is calculated as the number of true positives (as defined above) detected divided by the number of errors/reactions in the ground truth. 
    \item Number of false positives (\#FPs): This metric measures the number of false positives. 
    \item Overall F1 score (overall F1): This metric offers a balanced evaluation of detection and false positives. 
\end{itemize}

\subsection{Evaluation}
Challenge participants were given access to the training data and features to develop their ML models. Then, they were asked
to submit models and weights, and the organizers evaluated the submitted models on the test set (the test set
was released to the challenge participants without labels one month
prior to the submission deadline). To prevent overfitting on the test set, each participating group was allowed to submit their models and results for the test set up to three times. 

As participating teams may use different training and data processing techniques, we only considered the overall F1-score when comparing and ranking performance across teams. This metric simulates the model performance in a streaming setting and requires the model have a balance between detecting robot errors (true positives) and false positive. This provided additional flexibility in the choice of training mechanisms in the teams. Metrics were calculated using the same script provided to participants in the study repository. Challenge participants were also asked to submit a paper describing their model, and their works were peer-reviewed by two reviewers.

\subsection{Baseline Models}
We provided a multimodal baseline for each of the  sub-challenges\footnote{Baseline code and models: \href{https://github.com/ERR-HRI-Challenge/baseline2025}{https://github.com/ERR-HRI-Challenge/baseline2025}}. 

\subsubsection{Data Pre-Processing}
To facilitate training, we provided code that creates windowed chunks from the feature streams. A windowed chunk is considered to contain an error if it overlaps with the interval that is labeled robot error. We experimented with a range of window and step-sizes and decided on a window size of 3 seconds and step size of 0.5 seconds. Participating teams were encouraged to develop their own data pre-processing practices during models and were welcome to use different window sizes and step sizes. However, to prevent teams from using extremely large window sizes, which would result in significant delays in detection if these models were to be run in real-time, we limited the maximum allowed window size to 12 seconds for sub-challenge 1 and five seconds for sub-challenge 2. This maximum allowed window size is determined from the dataset and is set to be the average length of robot error/user reaction for that sub-challenge. 

We chose to use a window of three seconds with a step size of 0.5 seconds. We inputed any missing entries in the training features with the corresponding feature's mean. Since robot errors were sparse, this created a class imbalance, with only $11.35\%$ of the windows with robot errors for sub-challenge 1 and $3.33\%$ for sub-challenge 2.

\subsubsection{Training}
Due to the data imbalance between windows with robot errors and windows without robot errors, we applied synthetic minority over-sampling technique (SMOTE) to perform over-sampling to balance the training data. SMOTE synthesizes new minority-class examples in feature space to balance the classes, reducing bias toward the no-error class. 

We explored a set of standard modeling approaches---random forest (RF), explainable boosting machine (EBM), Extreme Gradient Boosting (XGBoost), long short-term memory (LSTM), and single-layer transformers---while leaving room for participants to innovate their approaches for detection and classification. We decided to use random forest as it performed best out of these during our exploration. 

We used leave-one-out cross validation and performed hyperparameter tuning (we tuned the number of estimators, max depth, min samples split, min samples leaf, max features, and class weight) using AUC on the validation set as the metric to pick the best hyperparameters. In the end, we trained a random forest using the best hyperparameters on all the training data and predicted on the held-out test set. In line with the process for challenge participants, we adjusted the model based on feedback on the model performance on the test set. We adjusted the predicted threshold (sub-challenge 1: 0.55 and sub-challenge 2: 0.45) to reduce the number of false positives in sub-challenge 1 and increase the number of true positives in sub-challenge 2. The resulting model had an overall F1 score of 0.24 in sub-challenge 1 and an F1 score of 0.29 in sub-challenge 2. The baseline models' performance on window-based evaluation metrics is shown in Table~\ref{tab:performance-window} and the models' performance on the \emph{on-the-fly} evaluation metrics is shown in Table~\ref{tab:performance-overall}

\begin{table}[]
\caption{Baseline models' window-based performance for the two sub-challenges.}
\begin{tabular}{c|cccc}
                & \textbf{AUC}  & \textbf{F1}   & \textbf{Acc} &
                \textbf{UAR} \\ \hline
\textbf{Sub-challenge 1} & 0.59 & 0.49 & 0.83 & 0.50   \\ \hline
\textbf{Sub-challenge 2} & 0.70  & 0.51 & 0.97 & 0.51
\end{tabular}
\label{tab:performance-window}
\end{table}

\begin{table}[]
\caption{Baseline models' \emph{on-the-fly} performance for the two sub-challenges.}
\begin{tabular}{c|ccc}
                & \textbf{\%Detected} & \textbf{\#FPs} & \textbf{Overall F1} \\ \hline
\textbf{Sub-challenge 1} & 39.54\%  & 163 & 0.24  \\ \hline
\textbf{Sub-challenge 2} & 29.03\%  & 97 & 0.13
\end{tabular}

\label{tab:performance-overall}
\end{table}

\subsection{Participation and Conclusion}
This paper introduces the ERR@HRI 2.0 Challenge organized in conjunction with the ACM International Conference on Multimedia 2025 (ACM-MM'25), which focuses on detecting robot errors and failures in human-robot conversations. A total of nine teams registered to participate in the challenge. Two teams submitted models to each of the sub-challenges and surpassed the baseline model performance. The two teams will be invited to submit a workshop-style paper describing their ML solutions and results on the dataset, as well as a publicly available code repository. We hope to continue hosting further ERR@HRI Grand Challenges in the coming years utilizing diverse human-robot interaction multimedia datasets. Future challenges should explore multimodal detection of different types of errors (physical, cognitive), severity of errors, types of robots (physical manipulator, mobile robots) in different settings (in-the-wild) to further advance multimodal detection of robot errors and failures.  

\begin{acks}
S. Cao, M. Stiber, A. Mahmood \& C. M. Huang's work
was supported by the National Science Foundation award \#2141335 and National Science Foundation award \#2143704. M. Spitale has been supported by PNRR-PE-AI FAIR project funded by the NextGeneration EU program. H. Gunes has been supported by the EPSRC project ARoEQ under grant ref. EP/R030782/1.
This paper has been proofread by a language model (AI), and the authors have read through the resulting content to ensure it accurately reflects the original intent. We would like to acknowledge Yifan Xu for their assistance with the baseline model training and Nithish Krishna Shreenevasan for their contribution to the data labeling. 
\end{acks}

\clearpage

\bibliographystyle{ACM-Reference-Format}
\balance
\bibliography{references}


\end{document}